%% file: icon2018.tex
\documentclass[11pt,a4paper]{article}
\usepackage[hyperref]{acl2018}
\usepackage{times}

\aclfinalcopy 

\usepackage{datatool, filecontents}

\usepackage{scalerel}

\usepackage{comment}
\usepackage{balance}
\usepackage{amsmath}
\usepackage{pgfplots}
\usepackage{tabularx}
\graphicspath{ {./} }

\DeclareMathOperator*{\argmax}{arg\,max}

\title{Improving Computer Generated Dialog with Auxiliary Loss Functions\\ and Custom Evaluation Metrics}

\author{
  Thomas Conley \\
  University of Colorado\\Colorado Springs\\
  1420 Austin Bluffs Pkwy\\
  Colorado Springs, CO, USA\\
  {\tt tconley@uccs.edu} \\\And
  Jack St. Clair \\
  Haverford College \\
  \\
  370 Lancaster Ave \\
  Haverford, PA, USA \\
  {\tt jrstclair@haverford.edu} \\\And
  Jugal Kalita \\
  University of Colorado\\Colorado Springs\\
  1420 Austin Bluffs Pkwy\\
  Colorado Springs, CO, USA\\
  {\tt jkalita@uccs.edu} 
  }

\usepackage{amssymb,amsmath,amsthm,enumitem}

\begin{document}

\setlength{\abovedisplayskip}{3pt}
\setlength{\belowdisplayskip}{3pt}

\setlength{\abovecaptionskip}{0em} 
\setlength{\belowcaptionskip}{0em} 
\setlist[itemize]{itemsep=.5em, topsep=.5em}

\maketitle

\begin{abstract}
Although people have the ability to engage in vapid dialogue without effort, this may not be a uniquely human trait.  Since the 1960's researchers have been trying to create agents that can generate artificial conversation.  These programs are commonly known as chatbots.  With increasing use of neural networks for dialog generation, some  conclude that this goal has been achieved.  This research joins the quest by creating a dialog generating Recurrent Neural Network (RNN) and by enhancing the ability of this network with auxiliary loss functions and a beam search.  Our custom loss functions achieve better cohesion and coherence by including calculations of Maximum Mutual Information (MMI) and entropy.  We demonstrate the effectiveness of this system by using a set of custom evaluation metrics inspired by an abundance of previous research and based on tried-and-true principles of Natural Language Processing.
\end{abstract}

\input{Sections/Introduction.tex}

\input{Sections/ProblemStatement.tex}
\input{Sections/Background.tex}

\input{Sections/LossFunctions.tex}

\input{Sections/Architecture.tex}

\input{Sections/EvaluationMetrics.tex}
\input{Sections/Results.tex}
\input{Sections/FutureWork.tex}


\balance
\bibliographystyle{aaai}
\bibliography{icon2018}
\end{document}

%% file: Sections/Introduction.tex
\section{Introduction}

Computer scientists have tried to build chatbots for a long time, starting from the initial attempt at building an artificial psycho-therapist called Eliza \cite{Weizenbaum1966}. Because of the nature of psychotherapy, even with its limited abilities, Eliza was able to impress the populace at large, in addition to the research community.  Eliza worked simply by pattern matching, and produced inane responses when pattern matching failed to produce a meaningful response. 

The frame-based architecture used by \cite{Bobrow1977} in the GUS system was the predominant approach to building dialog agents for several decades. Apple's SIRI and other digital assistants have used this architecture \cite{Bellegarda2013,Bellegarda2014,Jurafsky2018}.  Such speech-based conversation agents used a Partially Observable Markov Decision Process \cite{Sondik1971}  in a frame-based architecture, to maintain a system of beliefs and updated the system using Bayesian inference. They also  used reinforcement learning \cite{Sutton1998} as necessary. 

Recently, researchers have started building chatbots by training machine learning programs on transcripts of conversations. 
\citeauthor{Ritter2011} (\citeyear{Ritter2011}) presented a data-driven approach to generating responses to Twitter status posts, using statistical machine translation, treating a status post as a question and the response as its ``translation".  
Of late, researchers have  built chatbots using Artificial Neural Networks (ANN) or Deep Learning \cite{Cho2014,Sutskever2014}.  ANN-based Seq2Seq models have been used by many recent chatbots  \cite{Vinyals2015,Li2016a,Li2016b,Shao2017,Wu2018}. 

Although the Seq2Seq framework has shown good results in dialogue generation, we believe that the evaluation of the dialogues can be better measured. The research presented in this paper examines the role that various auxiliary loss functions play in the quality of generated dialog by RNNs when trained on several conversational corpora. Our contribution lies in the detailed analysis of generated dialogues, using custom metrics, as we change the auxiliary loss function. We believe that this is the first time such detailed analysis of automatically generated dialogs has been carried out. We use a simple RNN model for training the conversation agents since our primary focus is on auxiliary loss functions. We believe that this approach will have general applicability in other neural network architectures as well.

%% file: Sections/ProblemStatement.tex
\section{Problem Statement}

\newcommand{\Q} { \mathcal{Q}  }
\newcommand{\A} { \mathcal{A}  }
\newcommand{\D} { \mathcal{D}  }
\newcommand{\R} { \mathcal{R}}

We define a dialogue as the sequence of text elements $\D$ generated by the interaction between two agents $\Q$ and $\A$.  Text elements are a sequence of characters, $t\in\{c_1,c_2,...,c_i\}$, where $c_i$ is a character from used in the words of the conversation vocabulary.  Each elements $t_i$ is shown as $q_i$ or $a_i$ to distinguish outputs from agents $\Q$ and $\A$ respectively.  A conversation is seeded with an initial text element $q_1$, and $\A$ responds with a follow-up statement $a_1$. As shown in Equation \ref{eq:seq},
\begin{equation} \label{eq:seq}
 \mathcal{D} = \langle\langle q_1, a_1\rangle,\langle q_2, a_2\rangle,\dots,\langle q_i, a_i\rangle\rangle 
\end{equation}
\noindent the sequence grows with the continuous application of function $\R(t)$ as in Equations \ref{eq:r1} and \ref{eq:r2},
\begin{eqnarray}
a_i     &=& \R(q_i) \label{eq:r1}\\
q_{i+1} &=& \R(a_i) \label{eq:r2}
\end{eqnarray}
\noindent which show that each element of the conversation is generated from previous elements.  The function $\R(t)$ is a forward pass through an RNN using sequence $t_i$  as input and is followed by a beam search of the RNN output. We improve sequence generation and the function $\R(t)$ by incorporating auxiliary loss functions during the beam search.

A typical loss function in the context of classification, computes error by comparing predicted values with true values; the errors are propagated backward during training.  However, a Seq2Seq model trains on a series of sequences without labeled answers, that is, without any knowledge of what the truth is.  Instead, these models rely on minimizing the cross-entropy between the input and the raw network output. No output sequences are created during training. 

We present auxiliary loss functions which are applied after training during sequence generation by the beam search. Each path through the answer space represents a single possible choice for the final sequence.  The best answer among all possible paths is chosen by optimization of these loss function. 

Finally, we present simple evaluation metrics for determining the efficacy of our dialogue generation model. 

%% file: Sections/Background.tex
`
\section{Related Work}

Using Seq2Seq models for dialogue generation has become commonplace in  recent years. \citeauthor{Ritter2011} (\citeyear{Ritter2011}) were the first to use a  model used for Statistical Machine Translation (SMT) to generate responses to queries by training on a corpus of query-response pairs. \citeauthor{Sordoni2015} (\citeyear{Sordoni2015}) improved Ritter et al.'s work by re-scoring the output of the SMT-based response generation system with a Seq2Seq model that took context into account. 

\citeauthor{Vinyals2015} (\citeyear{Vinyals2015}) used an RNN-based model with a cross-entropy based auxiliary loss function and a greedy search at the output end. \citeauthor{Wen2015} (\citeyear{Wen2015}) used LSTMs for joint planning of sentences and surface realization by adding an extra cell to the standard LSTM architecture \cite{Hochreiter1997}, and using the cross-entropy loss. They produced sentence variations by sampling from sentence candidates. \citeauthor{Li2016a} (\citeyear{Li2016b}) used Maximum Mutual Information (MMI) as the objective function to produce diverse, interesting and appropriate responses. This objective function was not used in the training of the network, but to find the best among candidates produced by the model at the output, during generation of responses.  Our paper is substantially inspired by this work. 

\citeauthor{Li2016a} (\citeyear{Li2016a}) applied deep reinforcement learning using policy gradient methods to  punish sequences that displayed certain unwanted  properties of conversation: lack of informativity, incoherence and  responding inanely. Lack of informativity was measured in terms of high semantic similarity between consecutive turns of the same agent. Semantic coherence was measured in terms of mutual information, and low values were  used to penalize ungrammatical or incoherent responses. 

\citeauthor{Su2018} (\citeyear{Su2018}) use a hierarchical multi-layered decoding network to generate complex sentences. The layers are GRU-based \cite{Cho2014}, and each layer generates words associated with a specific Part-Of-Speech (POS) set. In particular, the first layer of the decoder generates nouns and pronouns; the second layer generates verbs, the third layer adjectives and adverbs; and the fourth layer, words belonging to other POSes. They also use  a technique called teacher forcing \cite{Williams1989} to train RNNs  using the output from the prior step as an input. 

Despite the relatively new methods that are being proposed for question answering and dialogue generation, the evaluation of the the generated text still relies on metrics like BLEU (Bilingual Evaluation Understudy) \cite{Papineni2002}, a metric that was designed for evaluation of SMT.
BLEU computes scores for  individual translated  sentences by comparing overlaps in terms of n-grams  with a set of good quality reference translations. 
These measurments alone are insufficient for evaluating the effectiveness of dialogue generation systems.
 
\citeauthor{Li2016a} (\citeyear{Li2016a}) used two additional computable metrics: the length of the dialogue generated, and diversity of distinct unigrams and bigrams.  While this simple measure may be a good addition to BLEU we, believe that a wider set of evaluation metrics is needed.  Coh-Metrix \cite{Graesser2004} is a Web-based tool that  analyzes texts with over 100 measures of cohesion, language complexity, and readability. We have used Coh-Metrix extensively in the evaluation of dialogue from this research and it has provided a rich understanding of the quality of our results.

%% file: Sections/LossFunctions.tex
\section{Loss Functions}\label{sectionLossFunction}
Our training model employs a softmax cross entropy loss function for back-propagation during training.  Rather than modify this primary loss function, we concentrate on the auxiliary loss function needed during sentence generation.  This function operates on partially generated sequences during a beam search and is used to find consensus among a number of possible choices equal to the beam width.  We have tested extensively using a beam width of 2 since our functions are configured to process 2 parameters.  We leave the expansion of this process to handle wider beam widths as an obvious future enhancement. 

We begin our testing using no auxiliary loss function at all and rely on network predictions alone to select subsequent characters.  We call this Network Loss (NET) in this research and consider the results a control baseline for comparison with other functions.

We continue testing with a basic MMI loss function $\hat{T}_{MMI}$ as shown in Equation \ref{eq:mmi}, where $S$ represents the current set of solution states during sentence generation and $T$ represents the set of possible next states.  This function is modeled after work conducted by \cite{Li2016b}.  The weighting factor $\lambda$ is configurable at run time and is used to adjust the relevance of current solution states versus future solution states, in the decision process.
\begin{equation}\label{eq:mmi}
\scaleto{c}{1pt}
\hat{T}_{\scaleto{MMI}{3.8pt}}  =\argmax_{T}\big\{\log p(T|S)-\lambda\log p(T)\big\}
\end{equation}

The basic MMI approach is suggested by \cite{Estevez2009} and implemented as shown in Equation \ref{eq:mmi}.  We further develop this MMI approach by including Entropy normalization, as inspired by \cite{Trinh2018} by who used normalized MMI for feature selection. We calculate entropy from predicted network probabilities as shown in Equations \ref{eq:hp} and \ref{eq:mmient2}.

\begin{equation}\label{eq:hp}
H_S = \sum_{t=0}^{|S|} -P(S_t) \times log(P(S_t))
\end{equation}
\begin{equation}\label{eq:mmient2}
H_T = \sum_{t=0}^{|T|} -P(T_t) \times log((T_t)
\end{equation}
\noindent The minimum of these values is used to normalize our MMI value as in Equation \ref{eq:norm}.
\begin{equation}\label{eq:norm}
\hat{T}_{\scaleto{NORM}{3.8pt}} = \frac{\hat{T}_{\scaleto{MMI}{3.8pt}}}{min(H_S,H_T)}
\end{equation}
Finally we experiment with MMI entropy normalization where entropy is not calculated but measured directly from the training corpus in terms of character frequencies. Optimizing based on this function should affect the uniqueness of generated sentences.

%% file: Sections/Architecture.tex
\section{Architecture} \label{sectionArchitecture}
The core of our model is a stack of dense layers comprised of gated recurrent unit (GRUs) cells. We tested extensively on a configuration with 3 layers, each divided into 3 blocks, where each block contained 2048 GRUs. This architecture is based on a prior implementation available on-line\footnote{https://github.com/pender/chatbot-rnn}.   

The GRU stack is initialized with the previous state ($s_{t-1}$) and the current character encoding ($x_t$) at each time step $t$ in the character sequence.  The GRU output ($Y_t$) and the weights from the final stack layer ($W_t$) are combined with a bias ($b$) to produce logits at time $t$.  We define logits as the raw output of the GRU stack which can be normalized and passed to a softmax function to produce probabilities.  In this scheme, we update the logits by applying weights and biases from the last GRU layer as shown in Equation \ref{eq:logits}. 
\begin{equation}\label{eq:logits}
Logits = ( Output \times Weights ) + Biases
\end{equation}
\noindent The logits are then passed to a loss function for back propagation within the GRU stack.   We do not limit or pad the length of the input sequence but perform back propagation through time (BBTT), relying on TensorFlow's default truncated back-propagation capabilities.
\noindent Note that, output sequences (${y_0, ...,  y_t}$) are not generated during the training phase where only the logits are used for back-propagation.  It is after training, during testing or dialogue generation, that the logits are converted to probability using softmax.  Finally, probabilities are converted to character sequences using a beam search.

\begin{figure}[ht]
\includegraphics[width=.9\columnwidth]{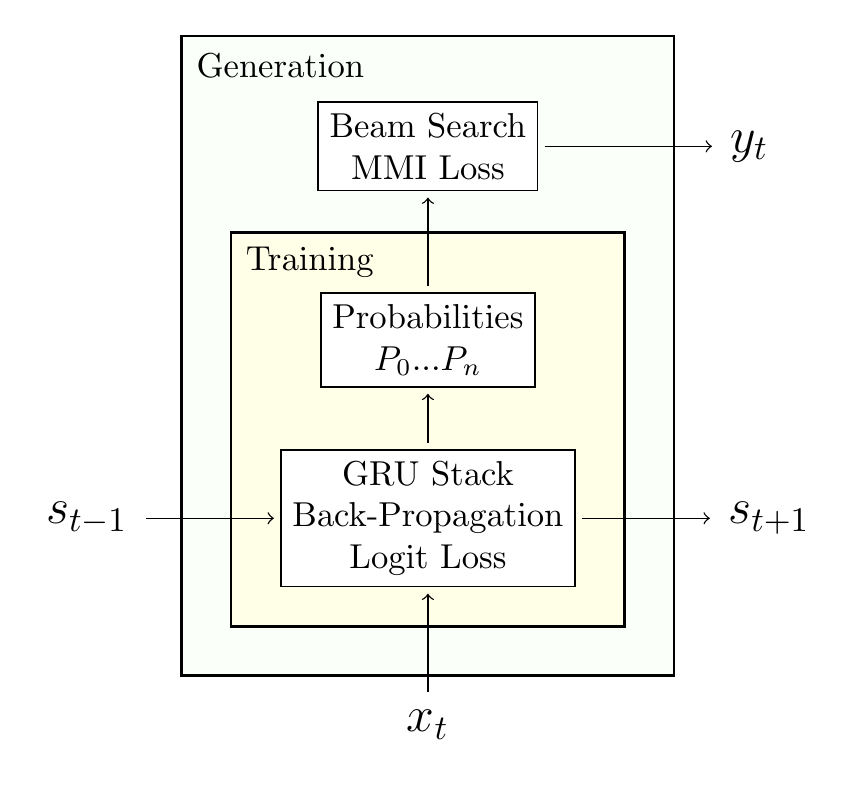}
\caption{Custom Loss Model}
\label{fig:model}
\end{figure}

Our beam search employs custom loss functions based on Maximum Mutual Information (MMI) as described in \cite{Li2016a}.  We extend this concept to include entropy-normalized MMI as discussed previously.
Figure \ref{fig:model} illustrates a single time-step $t$ in sequence processing by our recurrent neural network.

The model accepts a (one-hot) binary vector $X$ and a previous state vector, $S$, as inputs and produces a state vector, $S$ and a predicted probability distribution vector $P_t$, for the (one-hot) binary vector $Y_t$.

%% file: Sections/EvaluationMetrics.tex
\section{Evaluation Metrics}

Evaluation of generated text remains a difficult task as there is little consensus regarding what makes a good conversation \cite{liu2016not}. Word-overlap metrics such as BLEU \cite{Papineni2002}, METEOR \cite{Banerjee2005} and ROUGE \cite{Lin2004} have been used in the past, however, a simple overlapping of words between question and answer may not make for a good conversation and repetition may be considered annoying and reminiscent of Eliza, as mentioned earlier.

We begin our testing of generated dialogue using the on-line suite of tools provided by Coh-Metrix \cite{Graesser2004}.  Although this is a very manual process of cutting and pasting results, it provides insight from over 100 different metrics related to cohesion and coherence of text. 

After examining several of these measurements for effectiveness in evaluating our dialogues; we use the knowledge gained from this manual process to develop a few simple metrics that reflect the concepts of cohesion and coherence, but can be automated. We built our simple metrics using tried-and-true NLP standard modules such as WordNet \cite{Fellbaum1998}, GloVe \cite{Pennington2014}, NLTK \cite{Loper2002} and the Stanford CoreNLP \cite{Manning2014}. 

Inspired by the fore mentioned tools, we present four simple distance functions which we apply to sentences pairs from generated dialogues as a measure of coherence and cohesion. 
\begin{itemize}
\item {\em SynSet Distance} This metric uses a human generated semantic knowledge-base (WordNet) to create two sets of semantic elements, where elements consist of synonyms and lemmas evoked by the words of each sentence.  The ratio of the intersection of the sets to the union of the sets provides a distance measurement between 0 and 1.
\item {\em Embedding Distance} Here we exploit the semantic knowledge inherent in pretrained word embeddings to produce a set of the n-closest words from each word in a sentence.  Similar to {\em SynSet Distance} we use the ratio of the intersection of the sets to the union of the sets get a value between 0 and 1.
\item {\em Cosine Distance} We consider that the set of word embeddings from a sentence has semantic meaning in a manner similar to the well known concept of ``bag-of-words''. The cosine distance between the average of the two sets provides a result between 0 and 1.
\item {\em Sentiment Distance} A Naive Bayes Analyzer provides a simple measure of positive or negative sentiment, for each sentence.  With values between 0 and 1, a simple difference is used to represent {\em Sentiment Distance}.  
\end{itemize}

%% file: Sections/Results.tex
\section{Experiments and Results}
We tested our model by training on dialogue from Reddit and from the proceedings of the Supreme Court of the United States (SCOTUS) and by using four distinct auxiliary loss functions described in this research.  Network loss (NET), Maximum Mutual Information (MMI), Normalized MMI (NORM) and Entropy Normalized MMI (ENT) were used to generated conversations consisting of 15 question and answer pairs for testing. 

Using the Reddit trained model, multiple tests were run using Coh-Metrix and some results are summarized in Table \ref{tab:cohmetrix}. All test conversations consist of 15 question and answer pairs generated by two different chatbots. This summary of results provides insight into the relative effectiveness of our loss models as measured by Coh-Metrix. The definition of these metrics is left to \cite{Graesser2004}; however observed trends in Coh-Metrix have led to the development of our own custom metrics.

\setlength{\abovecaptionskip}{1em} 
\begin{table}[h]
\resizebox{\columnwidth}{!}{%
\begin{tabular}{lllll}
\hline
 & NET & MMI & NORM & ENT \\ \hline
Mean Words per Sentence & 10.070 & 3.200 & 1.550 & 51.389\\
Narrativity & 99.910 & 98.170 & 57.140 & 78.810\\
Syntactic Simplicity & 58.320 & 41.680 & 99.930 & 0.160\\
Referential Cohesion & 90.820 & 64.800 & 100 & 100\\
Sentence Semantic Similarity & 0.363 & 0.359 & 0.167 & 0.624\\
Lexical Diversity & 0.366 & 0.594 & 0.333 & 0.096\\
Connective Word Occurrence & 48.499 & 0 & 0 & 57.297\\
Modifiers per Noun Phrase & 0.408 & 0.231 & 0 & 0.908\\
Sentence Syntax Similarity & 0.114 & 0.158 & 0.593 & 0.040\\
Content Word Frequency & 2.813 & 4.580 & 2.358 & 2.835\\
Word Familiarity & 589.115 & 572 & 591.5 & 583.183\\
Reading Ease & 90.526 & 100 & 98.835 & 63.476\\
\end{tabular}%
}
\caption{Selected Coh-Metrix results from our model using four auxiliary loss functions.}
\label{tab:cohmetrix}
\vspace{-.5em}
\end{table}

\begin{figure}[ht]
\includegraphics[width=.9\columnwidth]{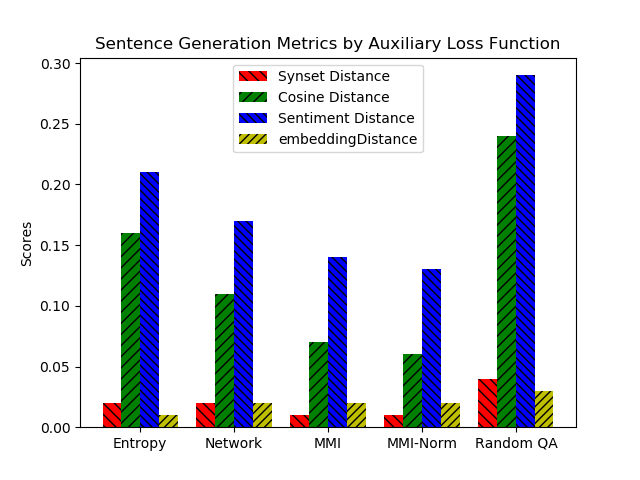}
\caption{Average distance measurements for custom auxiliary loss functions across all datasets.}
\label{fig:bars}
\vspace{-.5em}
\end{figure}

Comparative results shown in Figure \ref{fig:bars} indicate lower values for all 4 non-random metrics, showing that our system is not just parroting text sequences from the training corpus.  The larger results, produced by ENT, indicate that entropy normalization increases uniqueness in responses and thus increases the distance measure, as expected.  The lower measurements for the MMI based functions indicate a closer cohesion and coherence between question and answer; this may be a result of using lambda factor equal to .5 during testing which reduces the impact of previous solution states in favor of the predicted solution state.
 
\setlength{\abovecaptionskip}{0em} 


\subsection{Cohesion and Coherence}

A generated sample of text from SCOTUS, shown in Table \ref{tab:exampletext} illustrates the difference between cohesion and coherence. The fact that sentences seem to fit together well and flow naturally indicate high cohesion which can be produced by the neural network alone.  However, a close reading of the text shows that the network was unable to give logical sense to the words and sentences. The capitalization at the beginning of the sequence correctly shows name of a random speaker, as found in the training corpus. Our testing shows that a network built on a larger training set has greater cohesion dialogue of Table \ref{tab:exampletext} is reasonable, but no level of training alone  was able to create dialogue with any real logic or meaning.
\setlength{\abovecaptionskip}{1em} 
\begin{table}[h]
\small
\def\arraystretch{1.5} 
\begin{tabular}{ |p{.95\columnwidth}| }
	\hline "MR. COLE: I think we're talking about the district court to review it does, Your Honor. I believe that's correct, Justice Ginsburg. It's -- it's in navigation. If you have the distinction between aliens who we collect taxes. They're -- they're contested, would be able to read the restatement of the landowners -- or -- or that decision. In that instance, I think that was referred to the issue before this Court that have standing alone and then have set forth in these kinds of prosecutions, when i"\\
	\hline
\end{tabular}
\caption{Generated response from SCOTUS showing reasonable cohesion but a lack of coherence.}
\label{tab:exampletext}
\end{table}
\vspace{-1em}





%% file: Sections/FutureWork.tex
\section{Conclusion and Future Work}


Advancements in technology may allow development of more complex neural networks and more sophisticated loss functions.  With better evaluation models, a neural-network-based chatbot may be enhanced to learn more from itself using a better form of back-propagation, during the generation phase, as described in this research.  

Although human interaction is still considered to be the best method for dialog evaluation, future dialog generation models, based on this research, may be able to bring human level sophistication to computer generated text.

%% file: icon2018.bbl
\begin{thebibliography}{}

\bibitem[\protect\citeauthoryear{Banerjee and Lavie}{2005}]{Banerjee2005}
Banerjee, S., and Lavie, A.
\newblock 2005.
\newblock Meteor: An automatic metric for mt evaluation with improved
  correlation with human judgments.
\newblock In {\em Proceedings of the acl workshop on intrinsic and extrinsic
  evaluation measures for machine translation and/or summarization},  65--72.

\bibitem[\protect\citeauthoryear{Bellegarda}{2013}]{Bellegarda2013}
Bellegarda, J.~R.
\newblock 2013.
\newblock Natural language technology in mobile devices: Two grounding
  frameworks.
\newblock In {\em Mobile Speech and Advanced Natural Language Solutions}.
  Springer.
\newblock  185--196.

\bibitem[\protect\citeauthoryear{Bellegarda}{2014}]{Bellegarda2014}
Bellegarda, J.~R.
\newblock 2014.
\newblock Spoken language understanding for natural interaction: The siri
  experience.
\newblock In {\em Natural Interaction with Robots, Knowbots and Smartphones}.
  Springer.
\newblock  3--14.

\bibitem[\protect\citeauthoryear{Bobrow \bgroup et al\mbox.\egroup
  }{1977}]{Bobrow1977}
Bobrow, D.~G.; Kaplan, R.~M.; Kay, M.; Norman, D.~A.; Thompson, H.; and
  Winograd, T.
\newblock 1977.
\newblock Gus, a frame-driven dialog system.
\newblock {\em Artificial intelligence} 8(2):155--173.

\bibitem[\protect\citeauthoryear{Cho \bgroup et al\mbox.\egroup
  }{2014}]{Cho2014}
Cho, K.; van Merrienboer, B.; Gulcehre, C.; Bahdanau, D.; Bougares, F.;
  Schwenk, H.; and Bengio, Y.
\newblock 2014.
\newblock Learning phrase representations using rnn encoder--decoder for
  statistical machine translation.
\newblock In {\em Proceedings of the 2014 Conference on Empirical Methods in
  Natural Language Processing (EMNLP)},  1724--1734.

\bibitem[\protect\citeauthoryear{Est{\'e}vez \bgroup et al\mbox.\egroup
  }{2009}]{Estevez2009}
Est{\'e}vez, P.~A.; Tesmer, M.; Perez, C.~A.; and Zurada, J.~M.
\newblock 2009.
\newblock Normalized mutual information feature selection.
\newblock {\em IEEE Transactions on Neural Networks} 20(2):189--201.

\bibitem[\protect\citeauthoryear{Fellbaum}{1998}]{Fellbaum1998}
Fellbaum, C.
\newblock 1998.
\newblock {\em WordNet: An Electronic Lexical Database}.
\newblock Bradford Books.

\bibitem[\protect\citeauthoryear{Graesser \bgroup et al\mbox.\egroup
  }{2004}]{Graesser2004}
Graesser, A.~C.; McNamara, D.~S.; Louwerse, M.~M.; and Cai, Z.
\newblock 2004.
\newblock Coh-metrix: Analysis of text on cohesion and language.
\newblock {\em Behavior research methods, instruments, \& computers}
  36(2):193--202.

\bibitem[\protect\citeauthoryear{Hochreiter and
  Schmidhuber}{1997}]{Hochreiter1997}
Hochreiter, S., and Schmidhuber, J.
\newblock 1997.
\newblock Long short-term memory.
\newblock {\em Neural computation} 9(8):1735--1780.

\bibitem[\protect\citeauthoryear{Jurafsky and Martin}{2018}]{Jurafsky2018}
Jurafsky, D., and Martin, J.
\newblock 2018.
\newblock {\em Speech \& Language Processing (Third edition draft, available at
  https://web.stanford.edu/~jurafsky/slp3}.
\newblock Pearson.

\bibitem[\protect\citeauthoryear{Li \bgroup et al\mbox.\egroup
  }{2016a}]{Li2016b}
Li, J.; Galley, M.; Brockett, C.; Gao, J.; and Dolan, B.
\newblock 2016a.
\newblock A diversity-promoting objective function for neural conversation
  models.
\newblock In {\em Proceedings of the 2016 Conference of the North American
  Chapter of the Association for Computational Linguistics: Human Language
  Technologies},  110--119.

\bibitem[\protect\citeauthoryear{Li \bgroup et al\mbox.\egroup
  }{2016b}]{Li2016a}
Li, J.; Monroe, W.; Ritter, A.; Galley, M.; Gao, J.; and Jurafsky, D.
\newblock 2016b.
\newblock Deep reinforcement learning for dialogue generation.
\newblock {\em arXiv preprint arXiv:1606.01541}.

\bibitem[\protect\citeauthoryear{Lin}{2004}]{Lin2004}
Lin, C.-Y.
\newblock 2004.
\newblock Rouge: A package for automatic evaluation of summaries.
\newblock {\em Text Summarization Branches Out}.

\bibitem[\protect\citeauthoryear{Liu \bgroup et al\mbox.\egroup
  }{2016}]{liu2016not}
Liu, C.-W.; Lowe, R.; Serban, I.; Noseworthy, M.; Charlin, L.; and Pineau, J.
\newblock 2016.
\newblock How not to evaluate your dialogue system: An empirical study of
  unsupervised evaluation metrics for dialogue response generation.
\newblock In {\em Proceedings of the 2016 Conference on Empirical Methods in
  Natural Language Processing},  2122--2132.

\bibitem[\protect\citeauthoryear{Loper and Bird}{2002}]{Loper2002}
Loper, E., and Bird, S.
\newblock 2002.
\newblock Nltk: The natural language toolkit.
\newblock In {\em Proceedings of the ACL-02 Workshop on Effective Tools and
  Methodologies for Teaching Natural Language Processing and Computational
  Linguistics - Volume 1}, ETMTNLP '02,  63--70.
\newblock Stroudsburg, PA, USA: Association for Computational Linguistics.

\bibitem[\protect\citeauthoryear{Manning \bgroup et al\mbox.\egroup
  }{2014}]{Manning2014}
Manning, C.~D.; Surdeanu, M.; Bauer, J.; Finkel, J.; Bethard, S.~J.; and
  McClosky, D.
\newblock 2014.
\newblock The {Stanford} {CoreNLP} natural language processing toolkit.
\newblock In {\em Association for Computational Linguistics (ACL) System
  Demonstrations},  55--60.

\bibitem[\protect\citeauthoryear{Papineni \bgroup et al\mbox.\egroup
  }{2002}]{Papineni2002}
Papineni, K.; Roukos, S.; Ward, T.; and Zhu, W.-J.
\newblock 2002.
\newblock Bleu: a method for automatic evaluation of machine translation.
\newblock In {\em Proceedings of the 40th annual meeting on association for
  computational linguistics},  311--318.
\newblock Association for Computational Linguistics.

\bibitem[\protect\citeauthoryear{Pennington, Socher, and
  Manning}{2014}]{Pennington2014}
Pennington, J.; Socher, R.; and Manning, C.~D.
\newblock 2014.
\newblock Glove: Global vectors for word representation.
\newblock In {\em Empirical Methods in Natural Language Processing (EMNLP)},
  1532--1543.

\bibitem[\protect\citeauthoryear{Ritter, Cherry, and Dolan}{2011}]{Ritter2011}
Ritter, A.; Cherry, C.; and Dolan, W.~B.
\newblock 2011.
\newblock Data-driven response generation in social media.
\newblock In {\em Proceedings of the conference on empirical methods in natural
  language processing},  583--593.
\newblock Association for Computational Linguistics.

\bibitem[\protect\citeauthoryear{Shao \bgroup et al\mbox.\egroup
  }{2017}]{Shao2017}
Shao, Y.; Gouws, S.; Britz, D.; Goldie, A.; Strope, B.; and Kurzweil, R.
\newblock 2017.
\newblock Generating high-quality and informative conversation responses with
  sequence-to-sequence models.
\newblock In {\em Proceedings of the 2017 Conference on Empirical Methods in
  Natural Language Processing},  2210--2219.

\bibitem[\protect\citeauthoryear{Sondik}{1971}]{Sondik1971}
Sondik, E.~J.
\newblock 1971.
\newblock The optimal control of partially observable markov decision
  processes.
\newblock {\em PhD the sis, Stanford University}.

\bibitem[\protect\citeauthoryear{Sordoni \bgroup et al\mbox.\egroup
  }{2015}]{Sordoni2015}
Sordoni, A.; Galley, M.; Auli, M.; Brockett, C.; Ji, Y.; Mitchell, M.; Nie,
  J.-Y.; Gao, J.; and Dolan, B.
\newblock 2015.
\newblock A neural network approach to context-sensitive generation of
  conversational responses.
\newblock {\em arXiv preprint arXiv:1506.06714}.

\bibitem[\protect\citeauthoryear{Su \bgroup et al\mbox.\egroup }{2018}]{Su2018}
Su, S.-Y.; Lo, K.-L.; Yeh, Y.~T.; and Chen, Y.-N.
\newblock 2018.
\newblock Natural language generation by hierarchical decoding with linguistic
  patterns.
\newblock In {\em Proceedings of the 2018 Conference of the North American
  Chapter of the Association for Computational Linguistics: Human Language
  Technologies, Volume 2 (Short Papers)}, volume~2,  61--66.

\bibitem[\protect\citeauthoryear{Sutskever, Vinyals, and
  Le}{2014}]{Sutskever2014}
Sutskever, I.; Vinyals, O.; and Le, Q.~V.
\newblock 2014.
\newblock Sequence to sequence learning with neural networks.
\newblock In {\em Advances in Neural Information Processing Systems},
  3104--3112.

\bibitem[\protect\citeauthoryear{Sutton and Barto}{1998}]{Sutton1998}
Sutton, R.~S., and Barto, A.~G.
\newblock 1998.
\newblock {\em Introduction to reinforcement learning}, volume 135.
\newblock MIT press Cambridge.

\bibitem[\protect\citeauthoryear{Trinh \bgroup et al\mbox.\egroup
  }{2018}]{Trinh2018}
Trinh, T.~H.; Dai, A.~M.; Luong, T.; and Le, Q.~V.
\newblock 2018.
\newblock Learning longer-term dependencies in rnns with auxiliary losses.
\newblock {\em CoRR} abs/1803.00144.

\bibitem[\protect\citeauthoryear{Vinyals and Le}{2015}]{Vinyals2015}
Vinyals, O., and Le, Q.
\newblock 2015.
\newblock A neural conversational model.
\newblock {\em arXiv preprint arXiv:1506.05869}.

\bibitem[\protect\citeauthoryear{Weizenbaum}{1966}]{Weizenbaum1966}
Weizenbaum, J.
\newblock 1966.
\newblock {ELIZA} \textendash\ a computer program for the study of natural
  language communication between man and machine.
\newblock {\em Communications of the ACM} 9(1):36--45.

\bibitem[\protect\citeauthoryear{Wen \bgroup et al\mbox.\egroup
  }{2015}]{Wen2015}
Wen, T.-H.; Gasic, M.; Mrksic, N.; Su, P.-H.; Vandyke, D.; and Young, S.
\newblock 2015.
\newblock Semantically conditioned lstm-based natural language generation for
  spoken dialogue systems.
\newblock {\em arXiv preprint arXiv:1508.01745}.

\bibitem[\protect\citeauthoryear{Williams and Zipser}{1989}]{Williams1989}
Williams, R.~J., and Zipser, D.
\newblock 1989.
\newblock A learning algorithm for continually running fully recurrent neural
  networks.
\newblock {\em Neural computation} 1(2):270--280.

\bibitem[\protect\citeauthoryear{Wu, Martinez, and Klyen}{2018}]{Wu2018}
Wu, X.; Martinez, A.; and Klyen, M.
\newblock 2018.
\newblock Dialog generation using multi-turn reasoning neural networks.
\newblock In {\em Proceedings of the 2018 Conference of the North American
  Chapter of the Association for Computational Linguistics: Human Language
  Technologies, Volume 1 (Long Papers)}, volume~1,  2049--2059.

\end{thebibliography}
